\documentclass{article}

\usepackage[final]{corl_2022}
\usepackage{subfiles}
\usepackage{multirow}
\usepackage{amsfonts}
\usepackage{amsmath}
\usepackage{graphicx}
\usepackage{wrapfig}
\usepackage{enumitem}
\hypersetup{
    colorlinks=true,
    linkcolor=orange,
    filecolor=magenta,      
    urlcolor=orange,
    citecolor=orange,
}

\usepackage{titlesec}
\usepackage{pdfx}
\usepackage{placeins}

\newcommand{\revfour}[1]{{\color{black}{#1}}}

\newcommand{\algname}{CARBS: Coordinated Acquisition with Reactive Bimanual Scooping}
\newcommand{\algabbr}{CARBS}

\title{Learning Bimanual Scooping Policies \\ for Food Acquisition}

\author{
 Jennifer Grannen*, Yilin Wu*, Suneel Belkhale, Dorsa Sadigh \\
Stanford University, Stanford, CA \\
  
  \texttt{\{jgrannen, yilinwu\}@stanford.edu}
}
\makeatletter
\def\thanks#1{\protected@xdef\@thanks{\@thanks
        \protect\footnotetext{#1}}}
\makeatother
\thanks{*Equal contribution. }

\begin{document}

\titlespacing\section{0pt}{8pt plus 2pt minus 2pt}{0pt plus 2pt minus 2pt}

\maketitle

%===============================================================================

\begin{abstract}
A robotic feeding system must be able to acquire a variety of foods. 
Prior bite acquisition works consider single-arm spoon scooping or fork skewering, which do not generalize to foods with complex geometries and deformabilities. 
For example, when acquiring a group of peas, skewering could smoosh the peas while scooping without a barrier could result in chasing the peas on the plate.
In order to acquire foods with such diverse properties, we propose \emph{stabilizing} food items during scooping using a second arm, for example, by pushing peas against the spoon with a flat surface to prevent dispersion.
The added stabilizing arm can lead to new challenges.
Critically, this arm should stabilize the food scene without interfering with the acquisition motion, which is especially difficult for easily breakable high-risk food items like tofu. These high-risk foods can break between the pusher and spoon during scooping, which can lead to food waste falling out of the spoon.
We propose a general \emph{bimanual scooping primitive} and an \emph{adaptive stabilization strategy} that enables successful acquisition of a diverse set of food geometries and physical properties. 
Our approach, \algname{}, learns to stabilize without impeding task progress by identifying high-risk foods and robustly scooping them using closed-loop visual feedback. 
We find that \algabbr{} is able to generalize across food shape, size, and deformability and is additionally able to manipulate multiple food items simultaneously. \algabbr{} achieves 87.0\% success on scooping rigid foods, which is 25.8\% more successful than a single-arm baseline, and reduces food breakage by 16.2\% compared to an analytical baseline. Videos can be found on our \href{https://sites.google.com/view/bimanualscoop-corl22/home}{website}.

\end{abstract}

% TLDR summary: We develop a system for bimanual scooping that recognizes and prevents scooping failures to avoid breaking fragile food items during acquisition.

%===============================================================================

\section{Introduction}
\label{sec:introduction}

Approximately one million people in the U.S. depend on a caregiver's assistance to eat~\cite{brault2010americans}
, which can lead to malnutrition~\citep{buys2013spoons,Westergren2001eating} and an erosion of self-worth~\citep{perry2008assisted}. Building a robotic feeding system would enable patients to eat food independently~\citep{stanger1994devices}. A key component of such an assistive feeding system is bite acquisition, i.e., the act of a robotic arm picking up morsels of food from a plate for the goal of transferring the food to person's mouth~\citep{gallenberger2019transfer}.
Prior strategies for bite acquisition acquire food using a single robotic arm for either skewering with a fork or scooping with a spoon.
% motion primitives.
% a small set of utensils and hand-tuned motion primitives per utensil.
Fork-based bite acquisition learns to select skewering primitive parameters from a large supervised dataset of food items~\cite{tapo2019towards, gordon2020adaptive,feng2019robot}. 
However, fork-based skewering is inherently limited in what foods it can acquire. For example, a fork cannot skewer brittle cashews or small peas without damaging the food item, while scooping with a spoon might be more successful. 
% prior spoon scooping work is limited in generalizing to varied foods
Existing single-arm spoon-scooping bite acquisition often uses a hard-coded single-arm scooping primitive, making generalization to varied food items difficult~\cite{ohshima2012meal, park2020active}. When scooping more diverse foods, such as large, rigid-body fruit cubes and deformable cottage cheese, prior works rely on hard-coded adaptation strategies -- new primitives and even new tools -- which are not scalable. 
Similar to fork skewering, the single-arm scooping strategy is also inherently limited. For unstable items like broccoli or blueberries, it can be hard to know where exactly to scoop without pushing it off a barrier such as a fork or bowl wall. 
% Existing bimanual scooping methods rely on using a bowl wall to help stabilize the food item -- prior bimanual scooping work only considers scooping fruit from a melon~\cite{ureche2018constraints}. However, restricting the environment to a bowl also limits the range of foods and environments where these methods can be used and cannot be assumed for a general-purpose food manipulation robot. For example, pastas, vegetables, and tofu are often served on a plate, especially as subparts of a larger meal. 

\begin{figure} %[!htbp]
\vspace{-0.5cm}
\centering
\includegraphics[width=\linewidth]{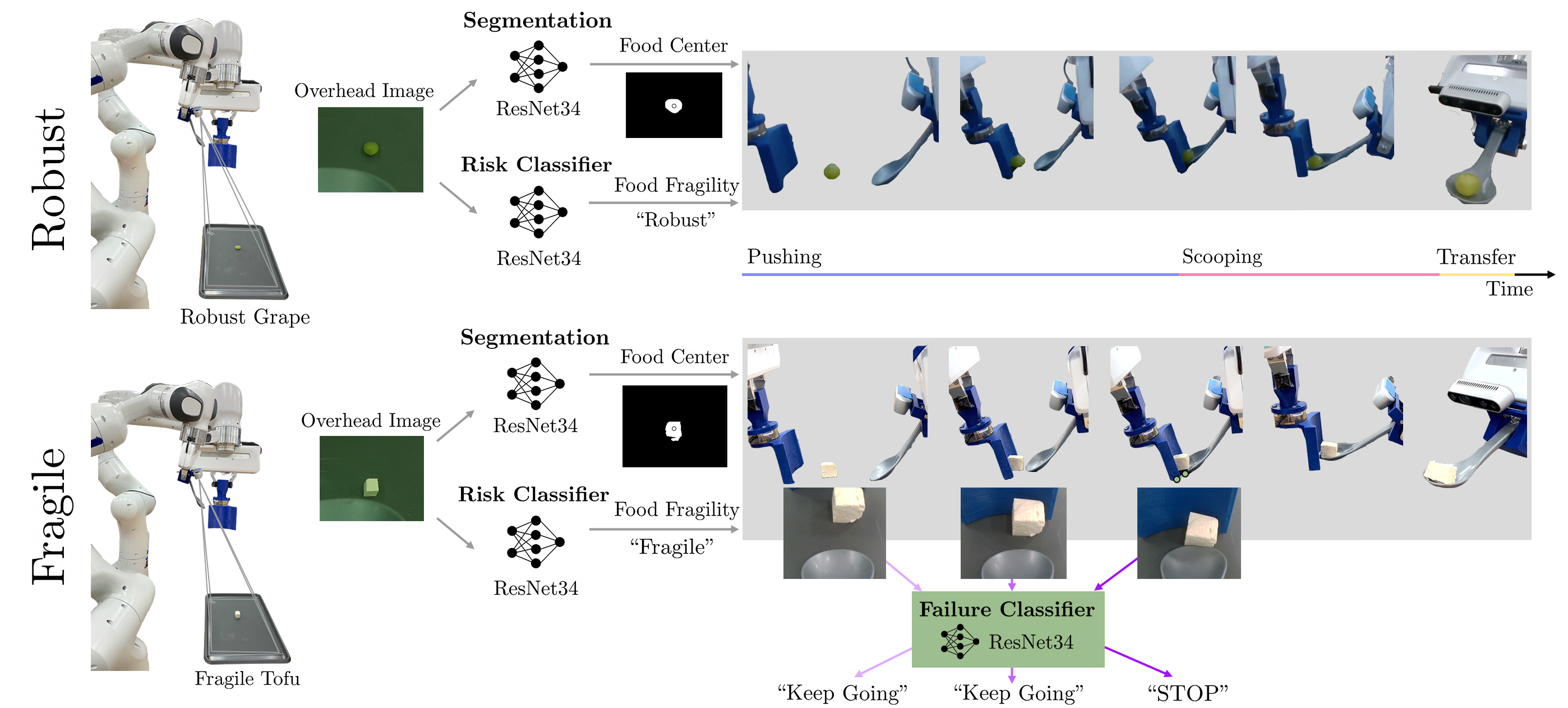}
\caption{\textbf{Learned Bimanual Scooping}: \algabbr{} is a bimanual scooping system for foods with varied geometries and deformabilities. \algabbr{} uses a second arm to stabilize food position during scooping. To avoid breaking deformable foods between the two arms, we learn a Risk Classifier and a Failure Classifier to identify high-risk, fragile foods and breakage-imminent states respectively.}
\label{fig:fig1}
\vspace{-0.8cm}
\end{figure}

\begin{wrapfigure}{l}{0.5\textwidth}
% \begin{center}
% \begin{minipage}[t]{0.5\linewidth}
% \begin{figure} %[!htbp]
\centering
\vspace{-0.2cm}
\includegraphics[width=\linewidth]{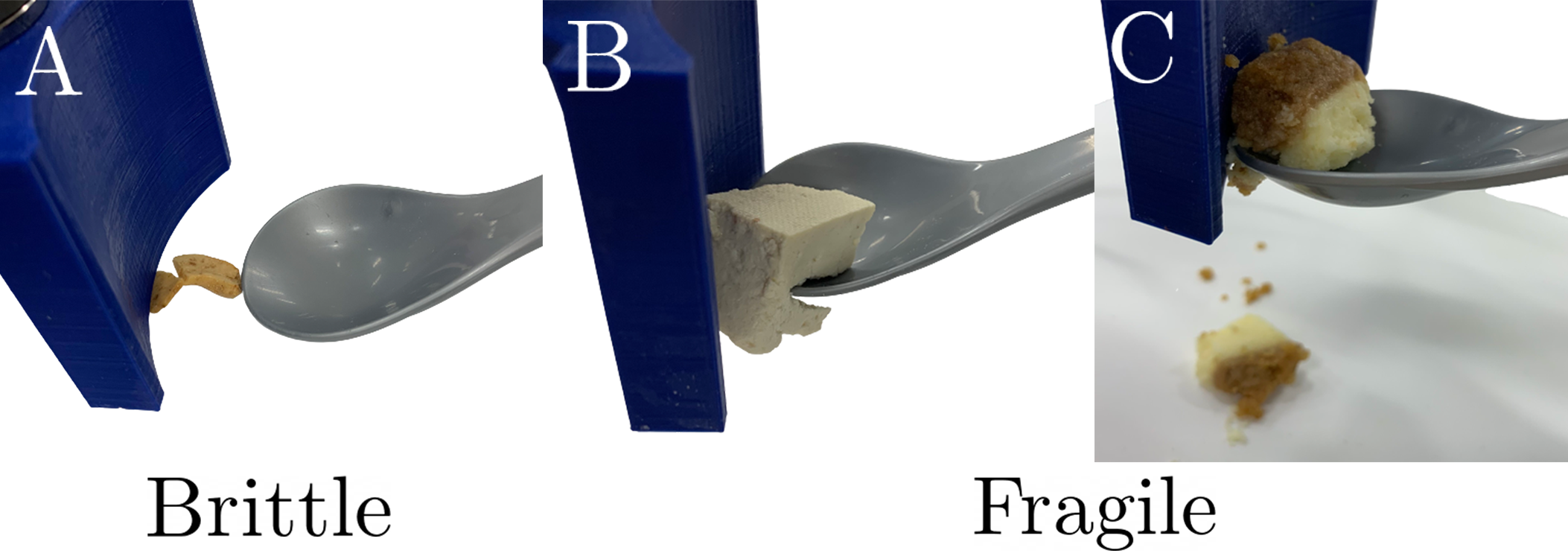}
\caption{\textbf{Bimanual Failures}: Adding a second stabilizing arm during scooping can lead to new breakage failures for foods of varying deformabilities. For both brittle and fragile food items like cashews (A), tofu (B) and cheesecake (C), food pieces can become wedged between the two arms and break due to excessive force being applied between the pusher and scooper. These breakages can cause food to fall out of the scooper (C), leading to food waste and scooping inefficiencies. }
\label{fig:bimanual_failures}
\vspace{-0.4cm}
% \end{figure}
% \end{minipage}
% \end{center}
\end{wrapfigure}

% prior work cannot scoop foods on a plate. what is hard about that?
% Scooping a variety of food objects on a plate presents new challenges.
% failures to motivate bimanual (geometries)
% It is even more difficult to scoop objects with unstable geometries or multiple objects without the help of a barrier -- for example, chasing peas on a plate is nontrivial and inefficient. 
Scooping objects with unstable geometries or scooping multiple objects without the help of a barrier -- for example, chasing peas on a plate -- is nontrivial and inefficient. 
% introduce bimanual contribution
To develop a generalizable scooping policy that can work across foods with difficult geometries, we need to use an additional robot arm to \emph{stabilize} the food item. For example, a primary arm should scoop peas against a surface that is pushing towards the spoon to stabilize the pea positions (See Fig.~\ref{fig:fig1}). As humans we intuitively and regularly use two arms to \emph{stabilize the environment} in many everyday tasks: when tying our shoes, we use a second arm to hold the knot; or similarly when cutting a steak we use a second arm to hold the steak to make the act of cutting easier for the first arm. As a result, there has been a growing focus on bimanual manipulation for tasks including rope untangling, peg insertion, food cutting, fabric manipulation, bottle opening, and bag opening, all implicitly leveraging the same insight that an additional arm can act in a stabilizing manner~\cite{zhang2019leveraging, grannen2020untangling, xu2022dextairity, shao2020learning, gemici2014learning, chitnis2020efficient}. In bite acquisition, we posit a second arm holding a \emph{pusher} (as shown in Fig.~\ref{fig:fig1}) can support a scooper in scooping objects with difficult geometries and deformabilities.

% failures to motivate learning (deformabilities)
However, adding a second arm for stabilizing the environment -- holding or pushing the food items -- opens the door to a new set of complications and failures (see Fig.~\ref{fig:bimanual_failures}): A pushing arm needs to physically make contact with the food to stabilize it. This can easily break or deform food items during scooping and thus impede on task progress. 
% Pushing foods into the scooper with a barrier can help with scooping some flat foods like cashews by building momentum to encourage the food into the scooper. However, this same motion can be hurtful for scooping unstable or round objects like blueberries as they may roll out of the scene. 
For objects with unstable geometries like snowpeas or macaroni, it is helpful for the barrier to follow the food into the scooper so the food does not fall out as the scooper finishes scooping, but this design comes at a cost. Fragile and deformable items like tofu or jello cubes can easily break when forced into the scooper by the pusher, which can also cause the item to fall off the scooper or leave residue on the plate.
% introduce learning contribution
As a result, it is difficult to define a single hard-coded primitive to that generalizes to both unstable geometry and breakage failures. 
We posit that many breakage-prone, or ``high-risk" foods will break under predictable scenarios when the pusher and scooper are squeezing the deformable food together. We employ this idea when scooping high-risk foods by detecting ``breakage-imminent" states and adjusting our scooping policy to anticipate and prevent food breakage and waste.
% We use the idea that many breakage-prone, or ``high-risk" foods have similar breakage failures. 
% To prevent these breakages, we propose learning two visual classifiers: the first to identify breakage-prone ``high-risk" foods and the second to recognize breakage-imminent scooping states. Given this feedback, we adjust our bimanual scooping primitive accordingly to avoid food breakage and waste. 

\emph{Our key insight is we need a second arm to effectively stabilize food environments by identifying high-risk food items, i.e. the ones that are breakable and fragile, and adapting a dynamic stabilizing strategy to anticipate and prevent such failures.}

% high level method overview, contribs
In this work, we propose a learned bimanual scooping policy, \algname{}, that uses a scooper arm and a pusher arm (similar to fork or knife) to acquire food items of a wide variety of properties, including shapes, sizes, and deformability. \algabbr{} first identifies \emph{high-risk food items} by classifying a visual observation of the target food item as fragile or robust. Next, \algabbr{} prevents breakage and dropping failures by \emph{servoing to adapt} the stabilizing action parameter: the distance between the scooper and pusher. This paper makes the following contributions:

\textbf{A Bimanual Scooping Primitive for Bite Acquisition.} To our knowledge, we are the first work to study bimanual strategies for acquiring widely varied foods. We define a novel bimanual scooping primitive and show that it generalizes to scooping 14 food classes with varied geometries and deformabilities. Our primitive is also the first to scoop multiple food items per acquisition action. 

\textbf{Learning to Avoid Bimanual Failures.} 
We contribute a framework for anticipating and preventing bimanual failures by identifying high-risk scenarios and adjusting the stabilizing parameter accordingly. To prevent breakage failures during bimanual scooping, this entails identifying deformable, fragile foods and adjusting the distance between arms during scooping. When scooping high-risk foods such as cheesecake and tofu, we learn to detect hazardous, breakage-imminent scooping states for closed-loop visual feedback to adjust our dynamic stabilizing policy. 
% We contribute a framework for anticipating and preventing scooping failures of deformable foods by adjusting the distance between arms during scooping. We first learn to identify high-risk foods, such as cheesecake and tofu. Then, we learn to detect hazardous and failure scooping states for closed-loop visual feedback to adjust our stabilizing scooping policy. 

\textbf{Evaluating Learned Scooping Policies.} We present physical experiments with \algabbr{}, which learns to identify fragile foods and adapt a stabilizing parameter to execute a bimanual scooping action. We find \algabbr{} is able to successfully generalize across food geometries and deformabilities to scoop 14 food classes without breakage in 85.7\% of trials. We will also open-source our food fragility and breakage datasets and the pusher and scooper CAD models.

\section{Related Work}
\label{sec:rw}

\noindent\textbf{Food Acquisition.}
Previous works have studied single-arm food acquisition with chopsticks~\cite{ke2021grasping, ke2020telemanipulation}, skewering with a fork~\cite{gallenberger2019transfer,tapo2019towards, gordon2020adaptive,feng2019robot}, and scooping with a spoon~\cite{ohshima2012meal, park2020active}.  
% \textcite{ke2021grasping} learn a manipulation policy from demonstrations to grasp a set of household objects with chopsticks, including potato chips. 
\citet{ke2021grasping} study grasping a set of household objects with chopsticks, but do not consider food objects or variations in geometric and physical properties. 
Past works on fork-based bite acquisition take a step towards more general food acquisition by learning an optimal skewering policy from a large dataset of food items~\cite{tapo2019towards, gordon2020adaptive,feng2019robot}. However, both chopsticks and forks with their accompanying acquisition primitives may struggle to generalize to many food items. In particular, it would be difficult to acquire fragile or very small foods such as jello or peas with a fork or chopsticks because the foods could break or would require very precise acquisition strategies unforgiving of slight errors. Spoon scooping is a promising acquisition alternative for these foods. \citet{ohshima2012meal} consider scooping portions off a large block of deformable foods -- tofu and pudding -- with an analytical policy. However, this analytical single-arm scooping method is limited to food items of relatively homogeneous geometries and deformability. \citet{park2020active} also use a single arm to scoop more varied foods (i.e. fruit cubes, cottage cheese) from a bowl by asking a human end-user to select between three spoon tools of differing materials and shapes per food class, which is not scalable to the large universe of potential foods. In contrast, our learned bimanual scooping approach uses one scooper and one pusher tool to acquire foods of highly variable geometries and deformabilities off a plate.

\noindent\textbf{Bimanual Manipulation.}
In recent years, bimanual manipulation has enabled robots to perform new tasks using new motion primitives such as bag opening, flinging fabrics, cutting vegetables, and opening bottles~\cite{grannen2020untangling, zhang2019leveraging, ha2021flingbot, kataoka2022bi, salehian2016coordinated, billard2022learning, chitnis2020efficient, gemici2014learning}. 
% In addition, bimanual manipulation can use handoffs to reach and operate in larger ranges of space compared to single-arm manipulation for tasks like object pick-and-place~\cite{shome2019anytime}.
The extra mobility provided by an additional arm can take the role of stabilizing the environment to reduce non-stationarity and make the task easier. For example, past works have implicitly considered utilizing a stabilizing arm for tasks including peg insertion, untangling ropes, and cutting foods~\cite{zhang2019leveraging, chitnis2020efficient, grannen2020untangling, shao2020learning, gemici2014learning}.
This idea is very relevant to the food manipulation domain. Foods can have widely varied physical properties and geometries -- they can be deformable, brittle, slippery, and in unstable shapes and poses. As a result, it can be difficult to manipulate these objects due to the unpredictable dynamics of food movements and interactions.

Past bimanual food manipulation works have considered food preparation tasks -- cutting and peeling vegetables, scooping out a melon, and mixing in a bowl.
Food cutting or peeling works use a stabilizing arm to hold the food in place during the cutting motions, but these stabilizing strategies are largely stationary or analytical~\cite{gemici2014learning, zhang2019leveraging} and do not make additional task progress~\cite{figueroa2017learning}.
On the other hand, \citet{ureche2018constraints} use a more versatile stabilizing strategy for melon scooping, zucchini peeling, and bowl mixing by learning bimanual interaction constraints from human demonstrations. For melon scooping, the stabilizing arm holding the melon is able to adjust its force to brace against the scooping tool to better stabilize the melon's position and even progress towards task success by pushing more melon into the scooper. However, this work is only specialized for scooping a melon and does not consider a generalizable policy for the large variety of food items necessary for a scalable robotic feeding system. 
% Additionally, this work relies on collected expert demonstrations which can be computationally expensive. 
% there is a high time and hardware cost for collecting human demonstrations. 
Our method similarly uses a dynamic stabilizing strategy that pushes food towards the scooper to both stabilize food position and make more task progress.
However, our method is able to learn to directly sense states close to bimanual constraint violations (food breakage failures) from visual feedback, and bypasses the need for expensive human demonstration collection. Our method is also able to learn these constraints in a food-agnostic way and generalizes to visually varied, out-of-distribution food classes. 
% \subfile{3-problem_statement.tex}
\section{\algname{}}
\label{sec:methods}

% move problem statement here
We consider the task of scooping a variety of food items off a plate while maximizing the integrity of the food item, i.e., the weight, after the scooping motion.
% without food damage or waste. Concretely, we want to successfully transfer the food from the plate to the scooper and also maximize the integrity of the food item, i.e., the weight, after the scooping motion. 
% Food items may vary in geometry and deformabilities. 
Food items may vary in geometry and deformabilities, including foods that are brittle (i.e. cashews), compliant (i.e. pasta), and fragile/breakable (i.e. tofu).
% For geometries, we consider variations in size and shape, including thicknesses, roundness, and symmetry. For deformabilities, we consider foods that are brittle (i.e. cashews), compliant (i.e. pasta), and fragile/breakable (i.e. tofu).
We assume access to a plate workspace with a standard $(x,y,z)$ coordinate frame (as in Fig.~\ref{fig:fig1}). 
We assume full bimanual access to this plate workspace with the following two mounted tools: \emph{Scooper} and \emph{Pusher} (See Fig.~\ref{fig:exp_setup}). The \emph{Scooper} tool is a plastic spoon mounted at an angle to the robot end effector with a camera mounted for access to angled images $I \in \mathbb{R}^{W \times H \times C}$ of the spoon and surrounding workspace. The \emph{Pusher} tool is a concave barrier that is mounted vertically to the robot end effector\footnote{This design is inspired by \href{https://www.etsy.com/market/food_pusher}{antique pushers} that were used to push foods into spoons.}. 

We model the bimanual scooping task as a \revfour{Partially Observable Markov Decision Process (POMDP) $\mathcal{M} = (\mathcal{O}, \mathcal{S}, \mathcal{A}, \mathcal{T}, \mathcal{R})$. 
We observe images $I \in \mathcal{O}$ of the unknown food environment states $S \in \mathcal{S}$} and define an action space $\mathcal{A}$ as joint 14-DoF robot actions $(a^s, a^p)$. The visual state space is an RGB image observation space $O \in \mathbb{R}^{W \times H \times C}$. We assume unknown transition dynamics $\mathcal{T}: \mathcal{S} \times \mathcal{A} \xrightarrow{} \mathcal{S}$, an initial state distribution of food configurations $\rho_0$, and a time horizon $T$. We define a reward $\mathcal{R}: \mathcal{S} \times \mathcal{A} \xrightarrow{} \mathbb{R}$ as the weight of the food in the scooper after scooping. We aim to construct a closed loop policy $\pi:\mathcal{O} \xrightarrow{} \mathcal{A}$ to maximize the expected reward by successfully scooping a set of varied foods.

\begin{figure} %[!htbp]
% \vspace{-0.8cm}
\centering
\includegraphics[width=\linewidth]{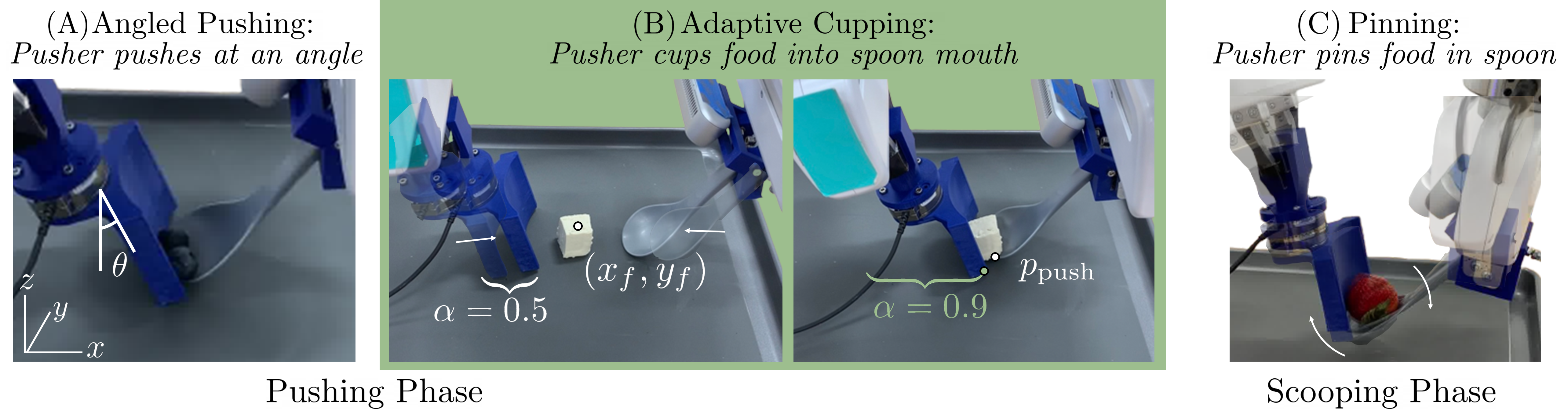}
\caption{\textbf{Bimanual Stabilizing Strategies}: \algabbr{} uses 3 stabilizing strategies: (1) \emph{Angled Pushing} and (2) \emph{Adaptive Cupping} during the Pushing Phase, and (3) \emph{Pinning} during the Scooping Phase. In \emph{Angled Pushing}, the pusher moves at an angle \revfour{$\theta = 15^{\circ}$} off the vertical, which encourages food items to roll off the barrier surface and into the spoon. In \emph{Adaptive Cupping}, the pusher pushes foods towards the spoon with the concave surface and cups them to be centered with the spoon mouth. This strategy is parameterized by a learned parameter $\alpha$ that represents the scaling of the distance travelled by the pusher. In \emph{Pinning}, the pusher moves upwards to follow the spoon mouth as it scoops to prevent food items from toppling and falling out. }
\label{fig:stabilizing_strats}
\vspace{-0.7cm}
\end{figure}

To scoop a large diversity of foods with two arms without dropping or breaking them, we introduce \algabbr{}, a reactive bimanual scooping policy learned from real food interactions (Fig.~\ref{fig:fig1}). We simplify the complex bimanual action space by introducing a novel bimanual primitive, which is parameterized by the \emph{distance of travel for the pusher} and the location of the food item, and employs three bimanual stabilizing strategies: (1) \emph{Angled Pusher}, (2) \emph{Cupping Motion}, and (3) \emph{Pinning Motion} (shown in Fig.\ref{fig:stabilizing_strats}). We show that this parameterization generalizes to robust food items (e.g., grape) by selecting a large distance of travel, as well as for breakage-prone items (e.g., tofu) which require more adaptive pusher travel distances. To handle the latter case of breakage-prone items, \algabbr{} learns to adapt the pusher travel distance by anticipating common types of failure. Our policy network learns to identify which regime we are in (robust or breakage-prone) based on just the initial plate image.

\noindent\textbf{A Bimanual Scooping Primitive.} We define a parameterized bimanual scooping primitive that takes two inputs, pusher travel distance $\alpha$ and food position $(x_f, y_f)$, and has three phases \revfour{to be performed in succession}: (1) \emph{Pushing}, (2) \emph{Scooping}, and (3) \emph{Food Transfer}. \revfour{We thus reduce the 14 DoF action space $\mathcal{A}$ to 3 dimensions: pusher travel distance $\alpha$ and food position $(x_f, y_f)$.} The scooper and pusher begin in starting positions centered around the food position along a fixed pushing axis. We empirically found that the choice of pushing axis did not affect performance, so we select the $x$-axis to favor our robots' range. 

During the \emph{Pushing} phase, both the pusher and scooper move towards each other along the $x$-axis towards some point $p_{\text{push}} = (x_f+d, y_f)$ closer to the scooper than pusher, as shown in Figure~\ref{fig:stabilizing_strats}\revfour{.A-B}. The Pushing phase also utilizes two bimanual stabilization strategies: \emph{Angled Pusher} and \emph{Cupping Motion} (See Fig.~\ref{fig:stabilizing_strats}\revfour{.A-B}). In Angled Pusher, the pusher is angled at some fixed angle \revfour{$\theta = 15^{\circ}$} about the $y$-axis, rather than orthogonal to the plate. This stabilizes the food position by encouraging the food to slide or roll into the scooper at the end of pushing when the two arms meet, as shown in Fig.~\ref{fig:stabilizing_strats}\revfour{.A}. In Cupping Motion, the pusher pins the food against the concave surface as it moves towards the scooper, which promotes centering the food position during movement into the entrance of the spoon (see Fig.~\ref{fig:stabilizing_strats}\revfour{.B}).
Cupping Motion is an adaptive stabilizing strategy that takes as input the primitive input $\alpha \in [0,1]$, which determines how close the pusher and scooper get to each other. $\alpha = 0$ implies no motion and $\alpha = 1$ results in the pusher and scooper reaching each other.
% We define a parameter $\alpha \in [0,1]$ for scaling the pushing distance traveled by the pusher. 
% When $\alpha = 1$, the pusher travels a distance of $d_p$ while the scooper travels $d_s$, and both arms meet at $p_{\text{push}}$. When $\alpha < 1$, the pusher travels $\alpha d_p$ and there is an additional $(1-\alpha) d_p$ distance left between the pusher and scooper. When $\alpha = 0$, both the scooper and pusher do not move, and directly terminate the Pushing phase.

After the Pushing phase, we assume \algabbr{} has successfully manipulated the food item into the scooper and move to the \emph{Scooping} phase. In this phase, we rotate the scooper up about the $y$ axis to ``scoop" the food into the bowl of the spoon. During the Scooping phase, \algabbr{} employs the \emph{Pinning Motion} stabilizing strategy where the pusher moves up along the $z$ axis with the scooper as it rotates (shown in Fig.~\ref{fig:stabilizing_strats}\revfour{.C}). This strategy prevents foods in unstable poses within the spoon from falling out of the scooper. Lastly, \algabbr{} finishes with the \emph{Food Transfer} phase by moving the pusher away from the scooper and rotating the scooper towards an end user to prepare for feeding.

% Our bimanual policy will learn to output the $\alpha$ and $(x_f, x_y)$ primitive inputs, as we describe next.

\noindent\textbf{Identifying High-Risk Settings.}
To learn the inputs of the bimanual scooping primitive $(x_f, y_f)$ and $\alpha$ \revfour{from image observations $I \in \mathcal{O}$}, \algabbr{} leverages the insight that foods with similar deformabilities encounter similar breakage failures, and posits that learning to identify high-risk settings, e.g. robust vs. breakage-prone foods, will help determine the optimal $\alpha$ inputs. \revfour{Due to the high variability of food dynamics, we do not learn a dynamics model of our POMDP $\mathcal{M}$ and instead learn parameters to perform adaptive scooping with our previously-defined bimanual scooping primitive, which performs each of the three phases in succession.}

Given an initial overhead image observation $I_0$, \algabbr{} starts by identifying the food position in the initial environment state $s_0$. To do this, we learn a segmentation model $f:\mathbb{R}^{W \times H \times C} \xrightarrow{} \mathbb{R}^{W \times H}$ to obtain a food mask and food center position $(x_f, y_f)$ from the initial image $I_0$, which is then passed to our bimanual scooping primitive. As illustrated in Fig.~\ref{fig:fig1}, to differentiate between food deformabilities, we learn a Risk Classifier $r:\mathbb{R}^{W \times H \times C} \xrightarrow{} \{0,1\}$ that identifies an initial food image $I_0$ as ``Robust" or ``Fragile". In practice, we instantiate the classifier with a ResNet34 model~\cite{he2016deep} trained on a hand-labelled food fragility dataset with 600 images of 14 food classes. 
For robust foods, we set $\alpha = 1$ for maximum pushing of the food, since the food is not in danger of breaking and can benefit from the added pushing stabilization. For fragile foods, it is nontrivial to select $\alpha$ given only an initial observation $I_0$, so we propose a closed loop system for determining $\alpha$.  

\noindent\textbf{Servoing for Fragile Foods.}
\revfour{At the beginning of each scooping rollout, we initialize $\alpha = 1$, indicating that the pusher should travel the full distance during the Pushing phase.} However, varied food dynamics and differences in deformability can dictate the need for different $\alpha$ values even within the same food class. For example, when scooping two pieces of tofu, the food items may deform or slide on the plate differently due to food interactions with the plate, slight robot imprecision, or food shelf life.
As a result, \algabbr{} uses closed loop visual feedback in the form of a Failure Classifier $f:\mathbb{R}^{W \times H \times C} \xrightarrow{} \{0,1\}$ that identifies breakage-imminent states where the food item is in contact with the pusher and scooper, but not yet squeezed until breakage. This classifier is run at each state during Pushing (as in Fig.~\ref{fig:fig1}). When a breakage-imminent state is detected, the Pushing phase is terminated and $\alpha < 1$. \revfour{For example, if a breakage-imminent state was detected after 65\% of the Pushing phase had completed, the phase would terminate early and this would correspond to $\alpha = 0.65$ because the pusher had traveled 65\% of the full pushing distance. }
We instantiate the Failure Classifier with a ResNet34~\cite{he2016deep} model trained on hand-labelled images of 30 scooping trials.

% summary section 
% In summary, \algabbr{} is able to scoop a variety of foods by defining a parameterized bimanual scooping primitive and learning the input parameters from closed-loop visual observations. 

% \subfile{4-carbs.tex}
\section{Experiments}
\label{sec:experiments}

\begin{figure} %[!htbp]
\centering
\vspace{-0.2cm}
\includegraphics[width=\linewidth]{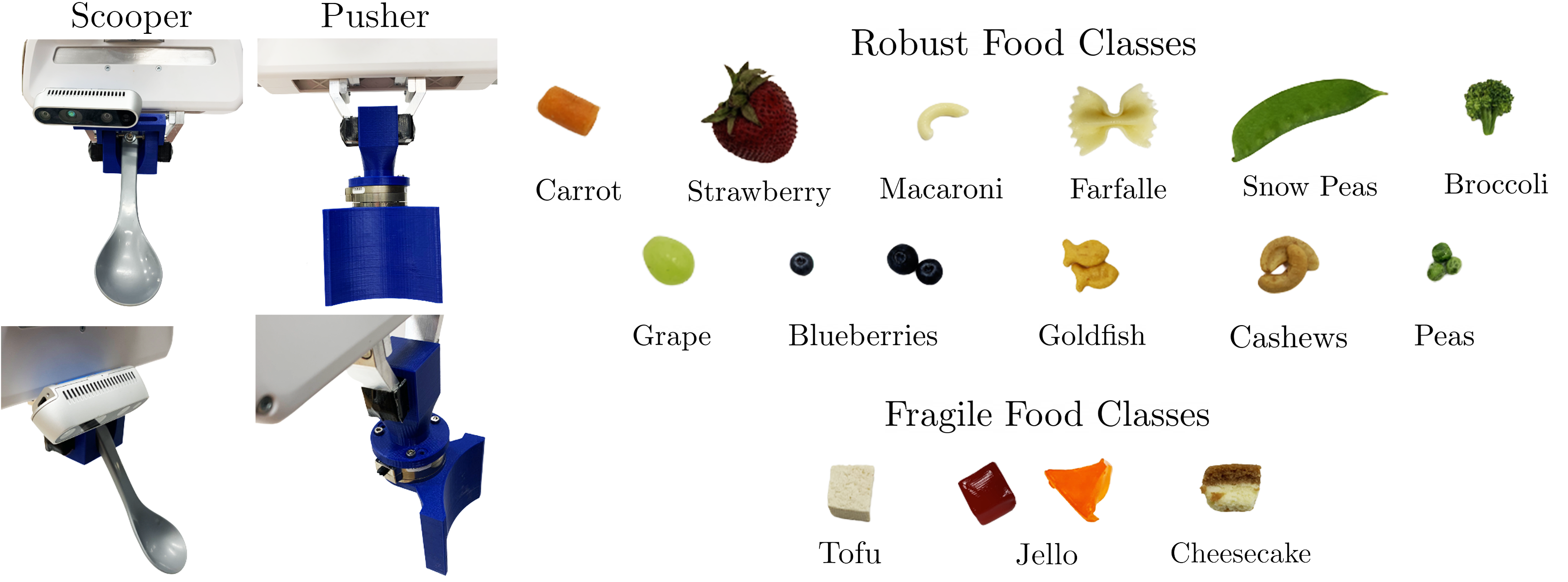}
\caption{\textbf{Experimental Setup}: \textbf{Left}: The Pusher is a custom concave barrier used to push food items to the mouth of the spoon. The Scooper has an RGB camera mounted above the spoon to access images of the spoon mouth and surrounding workspace during scooping. 
\textbf{Right}: We consider scooping 16 food settings: 11 robust containing up to three food items, and 4 fragile.}
\label{fig:exp_setup}
\vspace{-0.5cm}
% \vspace{-0.73cm}
\end{figure}

% For geometries, we consider variations in size148
% and shape, including thicknesses, roundness, and symmetry. For deformabilities, we consider foods149
% that are brittle (i.e. cashews), compliant (i.e. pasta), and fragile/breakable (
We validate \algabbr{}'s effectiveness on scooping foods of varied geometries and deformabilities. We design a series of experiments scooping 14 food items to demonstrate the advantage of a reactive bimanual strategy over hard-coded or single-arm actions. We select 14 food items to cover a wide range of sizes, shapes, and deformabilities: blueberry, broccoli, carrot, cashew, cheesecake cube, farfalle pasta, goldfish, grape, jello piece, macaroni pasta, pea, snow pea, strawberry and tofu cube (See Fig.~\ref{fig:exp_setup}). See Appendix \ref{sec:food} for food property details. We assume all food groups are pregrouped, where each food item is in contact with other items, if any, in the scene. 
Each set of food properties comes with a unique set of challenges. For example, blueberries, peas, and grapes are round, which may roll around the plate, while snowpeas and macaroni are irregularly shaped and can be difficult to stabilize within the spoon. We consider three deformable foods with diverse visual and material properties (jello, tofu, and cheesecake cubes), which are critically susceptible to breakage failures. 

\begin{wrapfigure}{r}{0.55\textwidth}
% \begin{table}[ht]
\centering
\vspace{-0.65cm}
% \vspace{-0.2cm}
% \resizebox{\columnwidth}{!}{
 \begin{tabular}{ p{0.29\linewidth}  p{0.1\linewidth} p{0.12\linewidth}  p{0.1\linewidth}   }

\multirow{2}{*}{\textbf{Food Type}} & \multicolumn{3}{c}{\textbf{Success Rate}}  \\
% \cline{2-4}
% \hline
% \multicolumn{1}{||c||}{} &  &   & A & B & C & D & E\\
 & \emph{Single} & $\alpha = 1$ & {\centering \emph{\algabbr{}}}  \\
% \multicolumn{2}{||c||}{}& \multicolumn{3}{c ||}{}& A & B & C & D & E\\
\hline
Broccoli & \textbf{5/5} & \textbf{5/5} & \textbf{5/5}  \\  
% \hline
Grapes & 3/5 & \textbf{5/5} & \textbf{5/5} \\  
% \hline
Blueberry & 4/5 & \textbf{5/5} & \textbf{5/5} \\ 
% \hline
Strawberry & \textbf{5/5} & \textbf{5/5} & \textbf{5/5} \\ 
% \hline
Carrot & 4/5 & \textbf{5/5} & \textbf{5/5}   \\ 
% \hline
Farfalle &  \textbf{5/5} & \textbf{5/5} & \textbf{5/5}  \\ 
% \hline
Macaroni & 2/5 & \textbf{5/5} & \textbf{5/5} \\ 
% \hline
Snow Pea & 3/5 & \textbf{4/5} & \textbf{4/5}  \\ 
% \hline
Cashews (2) & 2/10 & \textbf{7/10} & \textbf{7/10}  \\ 
% \hline
Goldfish (2) & 6/10 & \textbf{8/10} & \textbf{8/10}   \\ 
% \hline
Blueberries (2) & \textbf{6/10} & \textbf{6/10} & \textbf{6/10} \\ 
% \hline
Peas (3) & 3/15 & \textbf{15/15} & 14/15   \\ 

\end{tabular}
\makeatletter
\def\@captype{table}
\makeatother
\caption{\textbf{Robust Food Physical Results:}
\normalfont{ We report the per food item success rate over 5 trials of scooping robust foods with the Single, $\alpha = 1$, and \algabbr{} strategies. As expected, we observe \algabbr{} matches $\alpha = 1$ performance across all robust foods. 
Both the \algabbr{} and $\alpha = 1$ methods match or outperform the Single baseline, suggesting that the bimanual stabilizing strategies (Angled Pusher, Cupping Motion, and Pinning Motion) are advantageous over a static barrier. 
}}
% \vspace{-0.42cm}
\vspace{-0.55cm}
% \vspace{-0.3cm}
\label{table:robust_exps}
% \end{table}
\end{wrapfigure}

\noindent\textbf{Food Dataset.}
The Segmentation model is trained on a dataset of 600 overhead RGB images of all food items except Orange Jello in the real workspace. We subtract a background image of the workspace to obtain masks of the food items. 
The Risk Classifier is trained on a the same dataset of overhead RGB images of food items. We hand-label each image as ``Robust" or ``Fragile", and augment 8X. 
The Failure Classifier is trained on images from scooping rollouts of \emph{tofu only}, which we found to be sufficient for generalization to other food classes as well. More training details are in Appendix~\ref{sec:training}. We collect a dataset by recording 60 image frames each of the Pushing phase of 30 rollouts with $\alpha = 1$, meaning we push the food the maximum distance. We then hand-label when the food item breaks in each rollout, and automatically generate labels per image as ``Keep Going" or ``Stop" for safe and breakage-imminent states respectively.

\noindent\textbf{Implementation Details.}
Our real-world environment setup consists of two 7-DoF Franka Emika Panda arms, each holding a Scooper or Pusher tool as shown in Fig.~\ref{fig:exp_setup}. 
% where one gripper holds a custom 3D printed mount to control a plastic spoon and the other gripper holds a mounted concave pusher. 
The robot bases are set to be parallel with each other with the plate workspace between two bases\revfour{, and both robots are controlled with an impedance controller.} The Scooper is mounted at a 45 degree angle with an angled Intel Realsense D435 camera above the spoon as shown in Fig.~\ref{fig:exp_setup}. We designed a custom 3D printed pusher with a concave surface to encourage food grouping and stabilization. See Appendix~\ref{sec:appendix_hardware} for a discussion of design choices for the pusher and the spoon. \revfour{Our failure classifier to servo for breakage runs at 20 Hz frequency.}

 \begin{table*}[!htbp]
\centering
\vspace{-0.3cm}
% \resizebox{\columnwidth}{!}{
 \begin{tabular}{ c  c  c  c  c c  }

\multirow{2}{*}{\textbf{Food Type}} & \multicolumn{4}{c}{\textbf{Avg. Weight Difference (\%)}} & \multirow{2}{*}{$\alpha$ \textbf{Value}} \\
% \multicolumn{1}{c}{Stabilizing parameter (mm)} \\ 
\multicolumn{1}{c}{} & \emph{Single} & \revfour{$\alpha=0.93$} & $\alpha = 1$ & \emph{\algabbr{}} & \\
\hline
Tofu & 23.194 & \revfour{41.099} & 2.474 & \textbf{0.444} & 0.9477\revfour{$\pm 0.019$} \\ 
% \hline
Red Square Jello (Failure OOD) & 40.700 & \revfour{21.376} & 1.076 & \textbf{0.420} & 0.9400\revfour{$\pm 0.021$} \\ 
% \hline
Cheesecake (Failure OOD) & 26.873& \revfour{25.937} & 10.867 & \textbf{6.639} & 0.9231\revfour{$\pm 0.016$} \\ 
Orange Triangle Jello (\algabbr{} OOD) & 100 & \revfour{62.274}& 9.720&\textbf{0.449} & 0.9169\revfour{$\pm 0.030$} \\

\end{tabular}

\caption{\textbf{Fragile Food Physical Results:}
\normalfont{We report the weight loss of food items after scooping as a percentage of the original food weight, averaged across 5 trials. We also report the average values of \algabbr{}'s stabilizing parameter $\alpha$, a scaling value for the total pushing distance (13 cm) \revfour{with a 95\% confidence interval}. We observe that \algabbr{}'s Failure Classifier adjusts $\alpha$ to end pushing at a different position depending on the food. This suggests that the classifier learns to recognize the bimanual constraint that leads to breakage rather than a fixed ending position for all fragile foods. \revfour{We also find that the triangle jello has the highest $\alpha$ variability, possibly due to its irregular shape relative to the cube foods.} We observe \algabbr{}'s Failure Classifier generalizes to novel fragile food classes (Jello, Cheesecake) and its Risk Classifier generalizes to varied visual appearances and geometries within one class (Orange Jello). 
}}
\label{table:fragile_exps}
\vspace{-0.3cm}
 \end{table*}
\begin{wrapfigure}{r}{0.45\textwidth}
% \begin{center}
% \begin{minipage}[t]{0.5\linewidth}
% \begin{figure} %[!htbp]
\centering
\vspace{-0.2cm}
\includegraphics[width=0.95\linewidth]{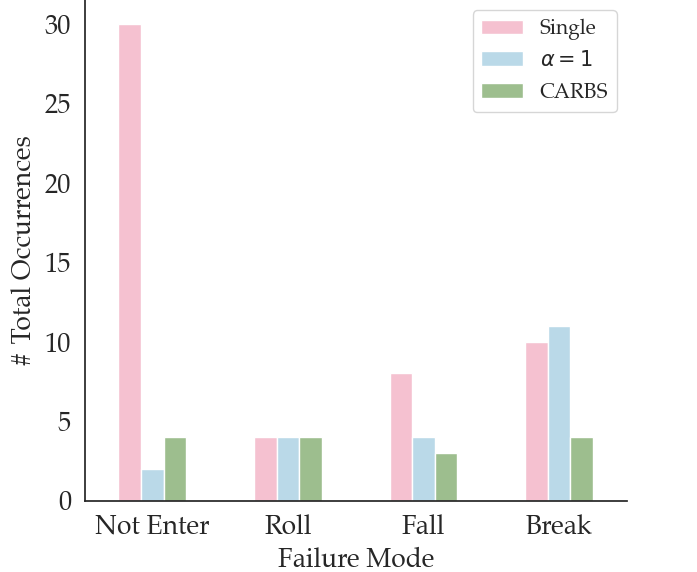}
\caption{\textbf{Failure Modes}:
We observe 4 failure modes across scooping strategies: (Not Enter) foods contact the scooper but do \emph{not enter} the spoon bowl, (Roll) foods \emph{roll out} of scooping range, (Fall) foods \emph{fall out} of the spoon after being scooped into the spoon, and (Break) \emph{breakage}. There are many (Not Enter) failures with the Single baseline because irregularly shaped foods and multiple items are difficult to roll into the spoon with a static pusher. We also find both baselines have a higher occurrence of breakage failures, supporting the need for an adaptive stabilizing strategy.}
\label{fig:failure_modes}
\vspace{-0.65cm}
% \end{figure}
% \end{minipage}
% \end{center}
\end{wrapfigure}

\noindent\textbf{Baselines.} \revfour{We compare against three baselines: \textbf{Single}, \boldmath{$\alpha = 0.93$} and \boldmath{$\alpha = 1$}}. \textbf{Single} executes a single-arm scooping method where the pusher is fixed and acts as a static barrier. The spoon moves towards the food and pushes the food against a stationary pusher during scooping. Notably, the pusher does not use any of the bimanual stabilizing strategies shown in Fig.~\ref{fig:stabilizing_strats}. \revfour{\boldmath{$\alpha=0.93$} executes a bimanual scooping primitive without an adaptive stabilizing strategy where the pushing distance is $0.93$ of the total 13cm pushing distance, stopping approximately 1cm early to prevent breakage. \boldmath{$\alpha = 1$} executes a bimanual scooping primitive instead with the full 13cm pushing distance ($\alpha = 1$)}, which is identical to the primitive for scooping a robust food item.
% \textbf{Single} and \textbf{Bimanual} that use different primitives for scooping and their motions are illustrated in Figure \todo{add a figure of three primitives maybe}
% \vspace{-0.25cm}
% \begin{itemize}[noitemsep,leftmargin=*]
    % \item \textbf{Single} executes a single-arm scooping method where the pusher is fixed and acts as a static barrier. The spoon moves towards the food and pusher and pushes the food against a stationary barrier during scooping. Notably, the pusher does not use any of the bimanual stabilizing strategies shown in Figure~\ref{fig:stabilizing_strats}.

    % \item \boldmath{$\alpha = 1$} executes a bimanual scooping primitive without an adaptive stabilizing strategy ($\alpha = 1$). Effectively, this baseline is identical to the primitive for scooping a robust food item.

% \end{itemize}

\noindent\textbf{Results.} We compare our method, \algabbr{}, against two baselines across 14 different food items, with 11 robust foods and 4 fragile foods. \revfour{We also include an additional $\alpha=0.93$ baseline for the 4 fragile foods, which is the average $\alpha$ value across all \algabbr{} fragile food trials.}
% robust foods
For robust foods, we report binary scooping success as whether the food ended within the spoon bowl after scooping in Table~\ref{table:robust_exps}. 
We consider settings of single foods of varied geometries, and additional settings of up to 3 food items. As expected, \algabbr{} and $\alpha = 1$ have similarly high performances because \algabbr{} should learn to set $\alpha = 1$ for Robust foods. We compare to the Single baseline to observe the advantages of the three bimanual scooping strategies (Angled Pushing, Cupping, and Pinning as in Fig.~\ref{fig:stabilizing_strats}) over a static pusher position.
% For single food items, we find that scooping foods that are round (i.e. grapes and blueberries) and foods with irregular geometries (i.e. macaroni and snow peas) benefit from the bimanual stabilizing motions. 
For round foods, the Angled Pusher stabilization (described in Sec.~\ref{sec:methods}) is important for building momentum and helping the item roll into the scooper, which prevents the food from rolling away (Roll Failure). For irregularly shaped foods that often extend past the spoon edges, the food items can be in unstable poses even once in the spoon bowl, which can cause them to fall out during the Scooping phase. These items benefit from the Pinning stabilizing strategy to pin the food in place throughout the rotation motion and prevent them from falling out (Fall Failure). \revfour{See Appendix~\ref{sec:ablation} for an ablation study on the stabilizing strategies.}

Lastly, we consider scooping 2-3 food items simultaneously, as inspired by the example of chasing peas around a plate. While \algabbr{} and the $\alpha = 1$ baseline still outperform the Single baseline, they fail to achieve as high success as scooping a single food item. It is more difficult to stabilize multiple food items simultaneously due to the added dynamics complexity. For example, stabilizing two blueberries to ensure \emph{both} roll into the mouth of the scooper is nontrivial (Not Enter Failure). The food dynamics also become more complicated as multiple foods can interact not only with the pusher and scooper tools, but also with each other. 

% A more in-depth analysis of failure modes is in Appendix \ref{sec:failure}.

% fragile foods
We present experiments scooping four fragile food settings in Table~\ref{table:fragile_exps}. Two food settings are out of distribution for our Failure Classifier and one is out of distribution for both the Failure and Risk Classifier. We report the weight loss during scooping as a percentage of the original food weight to measure the breakage failure severity. \algabbr{} is able to reduce food breakage by 16.185\% compared to the $\alpha = 1$ baseline. This suggests that the Failure Classifier can effectively recognize breakage-imminent states and adapt the stabilizing parameter $\alpha$ to prevent breakage. \revfour{We also find that $\alpha=0.93$ baseline, with a fixed early stop of 1cm to prevent breakage, has worse performance than \algabbr{} because the fixed distance cannot adapt to different food geometries and properties. Some smaller foods fail to fully enter the spoon bowl due to the early stop and are not scooped, resulting in 100\% weight loss. We note that although the average $\alpha$ value for cheesecake is 0.9231, there is still a large gap in performance between \algabbr{} and the \boldmath{$\alpha=0.93$} baseline because \algabbr{} is able to adjust $\alpha$ for each food item. We additionally report the 95\% confidence intervals for the \algabbr{} $\alpha$ values to highlight this adaptability.} We note that the weight difference for \algabbr{} for cheesecake, while still lower than the Single and $\alpha = 1$ baselines, is significantly higher than tofu and jello. This is due to the stickiness of the cheesecake and its propensity to leave food residue on the plate and tools during movement. We also report the $\alpha$ values learned with \algabbr{} and show that although our Failure Classifier is only trained on one food class (tofu), \algabbr{} is able to adjust $\alpha$ across novel food classes depending on their shape, size, and deformability. This supports our claim that learning to detect failures from vision generalizes across breakage-prone food classes. We also find \algabbr{}'s Risk Classifier generalizes within a food class to varied visual appearance and geometries, suggesting the effectiveness of \algabbr{} for scooping novel foods.

\section{Discussion}
\label{sec:discussion}

\textbf{Summary.} We present \algabbr{}, a learned bimanual scooping policy for robustly scooping food items of varied geometries and deformabilities. \algabbr{} learns a dynamic stablizing strategy to avoid breakage failures when scooping high-risk foods by identifying breakage-immenent states and adjusting the stabilizing action parameter: the distance between the scooper and pusher. We evaluate the generalizability of \algabbr{} with physical experiments scooping 14 foods of varying shapes, sizes, and fragility, and compare against two baselines. We find that \algabbr{} is able to successfully scoop 85.7\% of foods. 

\noindent\textbf{Limitations and Future Work.}
\algabbr{} struggles to scoop foods with uncommon material properties and complex dynamics, and multiple food items. While our system is able to reduce cheesecake breakage compared to baselines, it does not achieve similar success compared to other fragile foods due to the cheesecake's stickiness. \algabbr{} also struggles when scooping multiple blueberries with unpredictable dynamics. Their round shape and inertia allow them to roll off not only the scooper and pusher, but also each other. \algabbr{} leaves room for improvement when scooping multiple food items as well -- it is nontrivial to determine an optimal stabilizing policy for multiple items at once.
In future work, we hope to study more dynamic stabilizing strategies for food acquisition and other bimanual tasks, such as tying knots and buttoning clothes. We plan to relax our food environment assumptions and scoop an even larger range of foods, for example by pushing to group multiple scattered peas on a cluttered plate and then scooping into a spoon. \revfour{These cluttered food settings require longer horizon planning using potentially new primitives to group then acquire the food, which we leave to future work.} We will also explore multimodal sensing strategies \revfour{with new probing primitives for scooping} to generalize to more unseen foods \revfour{and augment our vision-only system}.

% Inspired by cheesecake's failure mode of sticking to the pusher and spoon, we aim to extend \algabbr{} to generalize to other food properties, including slipperiness and stickiness. We will study stabilizing strategies for more complex food dynamics, such as with multiple blueberries that can roll off not only the spoon and pusher, but off each other. Lastly, 

% challenging food dynamics, and out of distribution foods. We observe in physical experiments that cheesecake has relatively high weight loss during scooping compared to other foods due to its sticky nature. 
% something our method is bad here 
% - cheesecake is sticky so that's hard
% - blueberries roll a lot even with a pusher so that's hard. especially when there is more than one, they can roll off each other
% - we could collect more data or have a more intelligent way of learning what is high-risk vs not. haptic?

% something future work.
% \subfile{7-future_work.tex}

\clearpage
% \footnotesize
\acknowledgments{
This project was sponsored by NSF Awards 2132847, 2006388, and 1941722, and the Office of Naval Research (ONR). Jennifer Grannen is further grateful to be supported by an NSF GRFP. Any opinions, findings, conclusions or recommendations expressed in this material are those of the authors and do not necessarily reflect the views of the sponsors. We additionally thank our colleagues who provided helpful feedback and suggestions, especially Priya Sundaresan.}

\bibliography{t-refs}

\newpage
% \printbibliography

\newpage
\appendix
\begin{LARGE}
\begin{center}
\textbf{Learning Bimanual Scooping Policies\\ for Food Acquisition Supplementary Material}
\end{center}
\end{LARGE}
\maketitle
\label{sec:appendix}

This appendix contains food property details, classifier training details, experiment rollout visualizations, and a more in-depth analysis of failure modes. For more videos, please see our \href{https://sites.google.com/view/bimanualscoop-corl22/home}{website}.

\section{Food Property Analysis}
\label{sec:food}
% \todo{add food property table here (this should have fragile/robust classification as the first column)}
% - list the foods maybe and group into fragile or robust
% - some notable things to talk about:
%     - difference in jello consistency (the orange one is more/less stiff)
%     - difference in jello color
%     - difference in jello shape and how this affects the experiment results, specifically why was it so much hard for the bimanual and single baselines to scoop a triangle shape compared to a cube
The detailed food properties of 14 different types of food are presented in Table \ref{table:food_property}. Within the robust food category, we consider variations along brittleness and geometry. We also consider both brittle foods, e.g., cashews, and compliant foods, e.g., pasta. In addition, we experiment with a wide range of shapes, e.g., round grapes and irregularly shaped farfalle pasta. In our list, we also include foods with varied sizes, from extra small, e.g., peas, to larger items, e.g., strawberries, and we vary the thickness of food, e.g., thin snow peas vs thick broccoli. 

For fragile food, we use firm tofu, cheesecakes, and jello pieces of different colors, shapes and stiffness in the experiments. Unlike the tofu and jello pieces, cheesecakes have an additional property of stickiness that makes it extremely difficult to be scooped without any residue in the environment. For jello pieces, we have two colors and shapes of red square and orange triangle. The latter one is less stiff. One reason for the total failure of Single baseline on orange triangle jello is that the triangle is not symmetric along certain axis. Therefore, Single baseline is able to cup part of the orange triangle jello into the spoon mouth but it falls out of the spoon during the Scooping Phase because of the asymmetric weight imbalance.

% \begin{wrapfigure}{r}{0.55\textwidth}
\begin{table}[ht]
\centering
\caption{\textbf{Food Property Analysis:}
\normalfont{ We report the food properties of 14 different food classes in terms of deformability, brittleness and geometry, including shape, size and thickness. In the column of size, \textbf{L} represents large food, \textbf{M} represents medium food, \textbf{S} represents small food and \textbf{XS} represents extra small.
}}
% \vspace{-1.2cm}
% \resizebox{\columnwidth}{!}{
%  \begin{tabular}{ p{0.3\linewidth}  c{0.05\linewidth} c{0.05\linewidth}  c{0.05\linewidth}   }
\begin{tabular}{c | c | c | c  c c}
\multirow{2}{*}{\textbf{Food Type}} & \multirow{2}{*}{\textbf{Deformability}} & \multirow{2}{*}{\textbf{Brittleness}} & \multicolumn{3}{c}{\textbf{Geometry}}  \\
% \cline{2-4}
% \hline
% \multicolumn{1}{||c||}{} &  &   & A & B & C & D & E\\

 & & & Shape & Size & Thickness  \\
% \multicolumn{2}{||c||}{}& \multicolumn{3}{c ||}{}& A & B & C & D & E\\
 & & & & \\[-1em]
\hline
% \vspace{0.1cm}
 & & & & \\[-1em]
Broccoli & Robust & Compliant & Irregular & L & Thick \\  
% \hline
Grapes & Robust & Compliant & Round & M & Thick\\  
% \hline
Blueberry & Robust & Compliant & Round & S & Thick\\ 
% \hline
Strawberry & Robust & Compliant & Irregular & L & Thick\\ 
% \hline
Carrot & Robust & Compliant & Cylinder & M & Thick \\ 
% \hline
Farfalle &  Robust & Compliant & Irregular & M & Thin \\ 
% \hline
Macaroni & Robust & Compliant & Irregular & S & Thick\\ 
% \hline
Snow Pea & Robust & Compliant & Irregular & L & Thin\\ 
% \hline
Cashews & Robust & Brittle & Irregular & S & Thick\\ 
% \hline
Goldfish & Robust & Brittle & Irregular & S & Thick \\ 
% \hline
% \hline
Peas & Robust & Compliant & Round & XS & Thick \\ 
Tofu & Fragile & Compliant & Square & L & Thick\\
Cheesecake & Fragile & Compliant  & Square & L & Thick\\
Orange Triangle Jello & Fragile & Compliant & Triangle & L & Thick\\
Red Square Jello & Fragile & Compliant & Square & L & Thick\\

\end{tabular}
\makeatletter
\def\@captype{table}
\makeatother

% \vspace{-0.5cm}
\label{table:food_property}
\end{table}
% \end{wrapfigure}
\section{Training Details}
%  talk about data augmentation
%  talk about network architecture
%  talk about training and testing separation and the accuracy
\noindent\textbf{Risk Classifier}. We instantiate the Risk Classifier with a ResNet34 architecture trained on a hand-labelled dataset of 14 food classes labelled ``Robust`` or ``Fragile``. The dataset is composed of 600 overhead RGB images of all foods in Table~\ref{table:food_property} except Orange Triangle Jello and augmented by 8X by applying a series of standard label-preserving image transformations, including rotation, flipping, blurring, affine transformation, contrast changes, hue and saturation changes and addition of Gaussian noise. The applied augmentations not only enlarge the dataset but also enable Risk Classifier to work under various lighting conditions and be robust to small camera shift. With enough augmented data, we use Binary Cross-Entropy Loss and the Adam Optimizer \cite{kingma2014adam} with a learning rate of 1e-4 and a weight decay of 1e-4 to train the Risk Classifier. We train on a NVIDIA GeForce GTX 1070 GPU for 15 epochs. 

% - we train a 
% - @ Yilin: you should have the code for training that I shared with you for both of these

\noindent\textbf{Failure Classifier}. Similar to Risk Classifier, we also use the ResNet34 \cite{he2016deep} architecture to train a Failure Classifier to identify the breakage-imminent states during the pushing phase. The model is trained on the dataset of 30 rollouts, each containing 60 image frames, of the Pushing Phase of tofu with $\alpha = 1$ for maximum pushing distance. We augment the dataset by 8X with the same augmentations as with the Risk Classifier. We train the Failure Classifier with Binary Cross-Entropy Loss and the Adam optimizer \cite{kingma2014adam} with a learning rate of 1e-4 and a weight decay of 1e-4. We train on a NVIDIA GeForce GTX 1070 GPU for 25 epochs. \revfour{We find that the Failure Classifier is able to generalize to the three fragile food classes unseen during training with a success rate of 96.8\% of 157 images over 13 scooping rollouts.} 
\label{sec:training}

\begin{figure}
\vspace{-0.5cm}
\includegraphics[width=\linewidth]{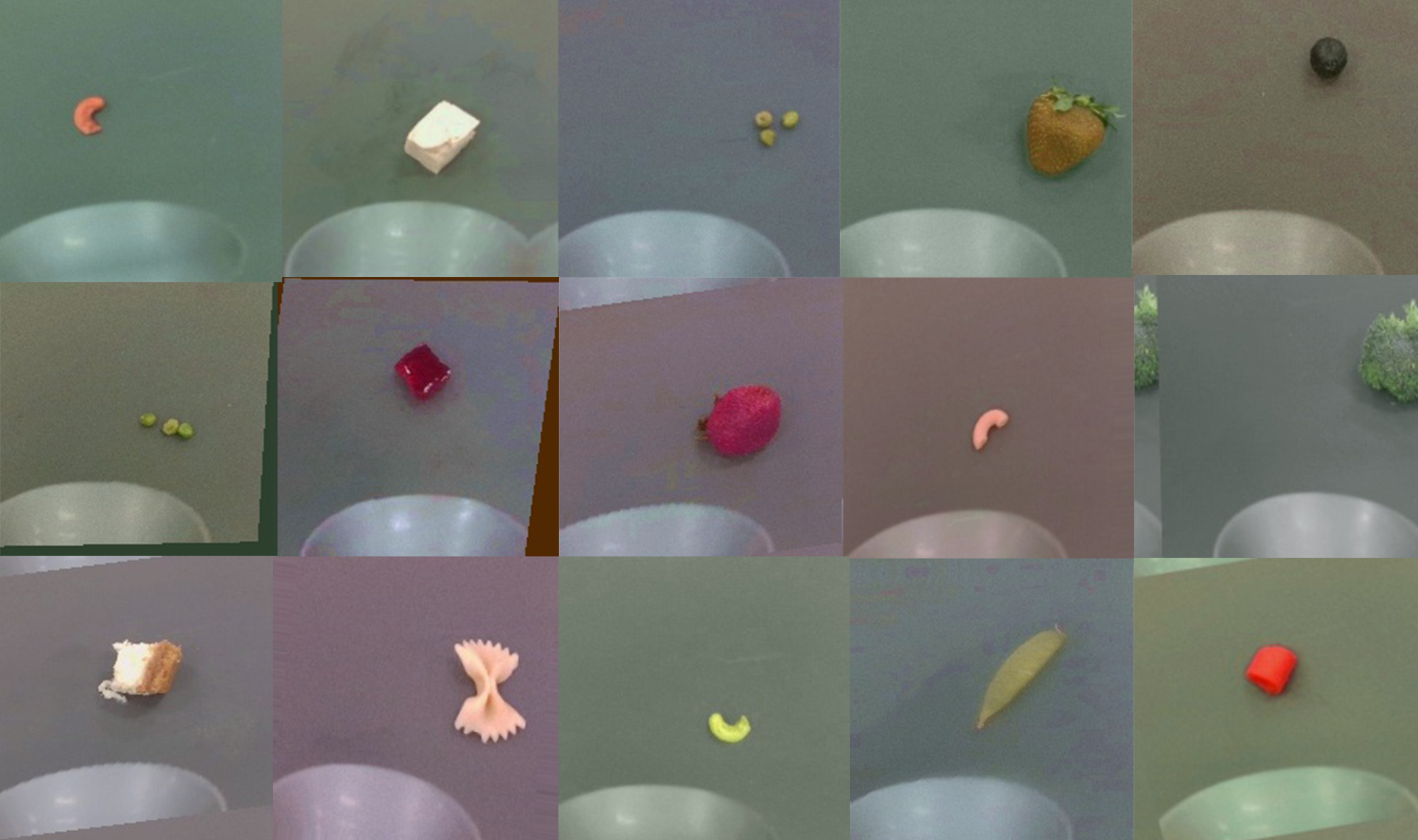}
\caption{\textbf{Augmented Dataset Images}: \revfour{We present examples of augmented overhead food images from the dataset used to train the Risk Classifier. The datasets used to train the Segmentation and Failure Classifier models were augmented with the same techniques.}}
\label{fig:data_aug}
\vspace{-0.3cm}
\end{figure}

 \begin{table*}[!htbp]
\centering
\vspace{-0.3cm}
% \resizebox{\columnwidth}{!}{
 \begin{tabular}{ c  c  }

\textbf{Augmentation} & \textbf{Parameters} \\ 
\hline
LinearContrast & (0.95, 1.05) \\ 
Add & (-10, 10) \\
GammaContrast & (0.95, 1.05) \\
GaussianBlur & (0.0, 0.6) \\
MultiplySaturation & (0.95, 1.05) \\
AdditiveGaussianNoise & (0, 3.1875) \\
Flipud & 0.5 \\

\end{tabular}

\caption{\textbf{Data Augmentation Parameters:}
\normalfont{\revfour{
We report the augmentation techniques used to train all models in \algabbr{}, along with their accompanying parameter values. All augmentations are used from the imgaug library~\cite{imgaug}.
}}}
\label{table:data_aug_params}
\vspace{-0.3cm}
 \end{table*}

\section{Hardware Design}
\label{sec:appendix_hardware}
% \todo{discuss the design choice of pusher and scooper}
% - we should talk about specific design details including:
% - why is the pusher concave
% - why is the pusher roughly the same size of the spoon
% - why is the spoon mounted at an angle, why this angle?
% - why is the camera mounted on the spoon

% generally talk about the design of the pusher and the mount of the spoon matters
% the design of the pusher is inspired by the antique pusher and why choose this size and this shape of concavity
% concave surface like a border to push the food to the center of the spoon
For our bimanual scooping task, we find that the design of the pusher and the scooper greatly improved the efficiency and effectiveness of the bimanual primitive.

Inspired by the antique pushers used by children to push food onto the spoon, we 3D print a custom concave pusher that has approximately the same curvature as the mouth of the spoon. During the Pushing Phase, the concave surface of the pusher pushes the off-centered food items to the mouth of the spoon and groups multiple food items together towards the center while a flat pusher may cause potential spreading of food over the plate. Another critical design choice is the pusher's size. It approximately the same size of the spoon so that the food items that are already grouped to the center of the pusher with the pusher's concavity. Foods are cupped into the spoon mouth without leaving anything beyond the reachable range of the spoon.

Apart from the design of the pusher, we also experiment with different designs of the scooper spoon tool and mount. The two key components of the scooper design are the mounted angle of the spoon and the tilted angle of the camera. The spoon is angled 45 degrees off the vertical axis, to align the mouth of the spoon to be tangent to the plate workspace and avoid robot motion range constraints. A larger angle would make the front part of the spoon mouth too high, which increases the difficulty of scooping thin foods, e.g., snow peas, and extra small foods, e.g., peas, because they can slide under the spoon mouth.
A smaller angle is prone to cause conflicts and break constraints of two arms during the Pushing and Scooping phase. The mounted camera is mounted at 30 degrees to capture the full view of the spoon mouth, the food in front of the spoon and the pusher during the Pushing Phase, which is necessary for the Failure Classifier.

\begin{figure} %[!htbp]
\centering
\vspace{-0.2cm}
\includegraphics[width=\linewidth]{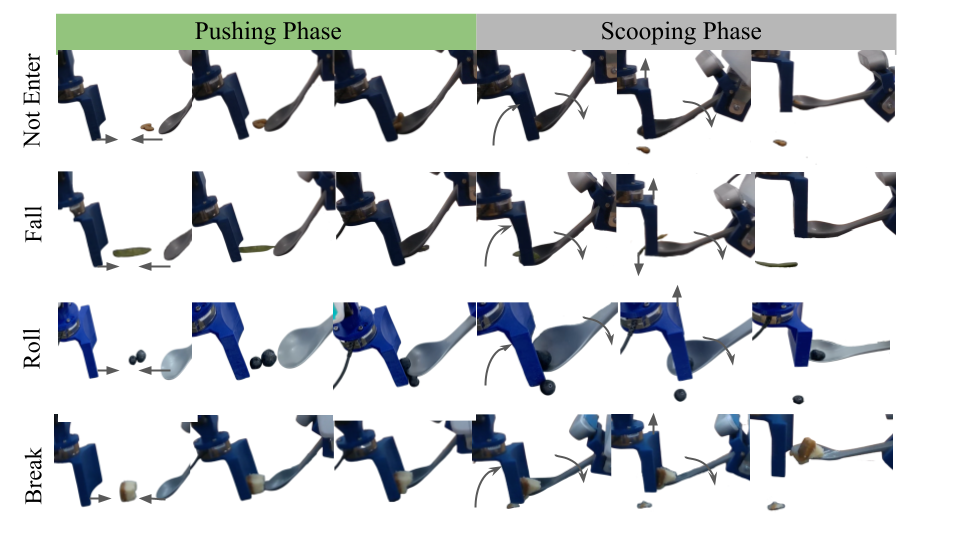}
\caption{\textbf{Failure Mode Rollouts}: We present selected rollouts illustrating each of the four failure modes. The Not Enter failure shows a cashew becoming wedged between the pusher and scooper and then failing to enter the spoon mouth.
The Fall failure shows the snow pea entering the spoon bowl, but falling out of the spoon off the side due to its irregular geometry.
The Roll failure shows two blueberries rolling off each other and out of the trajectory of the scooper.
Lastly, the final row shows a cheesecake piece breaking, leaving a piece of food on the plate after scooping.}
\label{fig:alpha1_baseline}
% \vspace{-0.73cm}
\end{figure}

\section{Failure Mode Analysis}
\begin{figure} %[!htbp]
\centering
\vspace{-0.2cm}
\includegraphics[width=\linewidth]{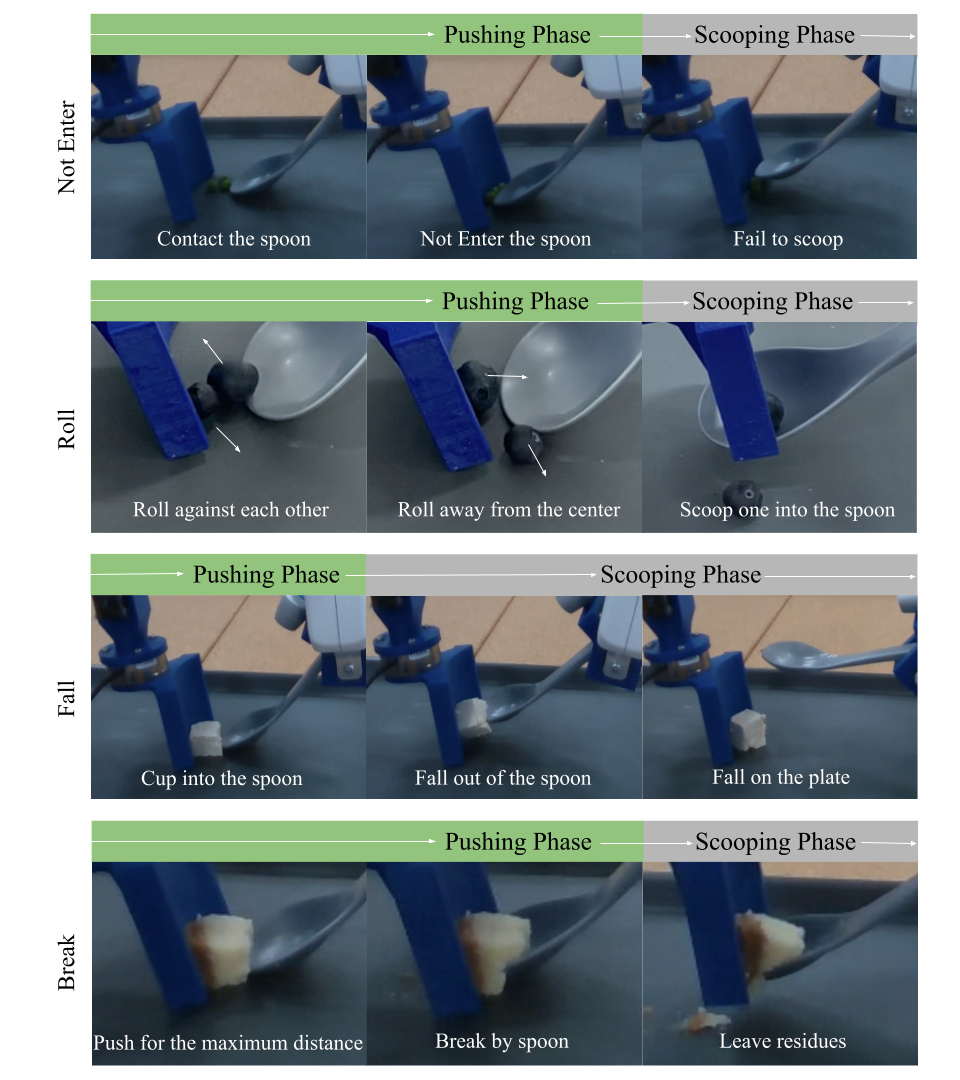}
\caption{\textbf{Failure Mode Visualizations:}: We present four failure modes: Not Enter, Roll, Fall and Break. Break failures are only present with deformable foods, such as cheesecake shown here. For the robust food category, Not Enter is the most common failure mode.}
\label{fig:failure_mode}
% \vspace{-0.73cm}
\end{figure}

% \todo{add rollout for each failure mode in each method and add more detailed analysis}
% - maybe a full rollout for each failure each a bit much, we can if we have time though! if you want to try making some figures with these, that would be good. we can also just do images of each.
% - but we should add more text describing each failure more in depth.
We observe four failure modes during experiments with all three bimanual scooping strategies: Not Enter, Roll, Fall, and Break. We present visualizations of these failure modes in the Fig.~\ref{fig:failure_mode}. 

\noindent\textbf{Not Enter}: As shown in Fig. \ref{fig:failure_modes}, the failure mode Not Enter is the most common failure mode for the Single baseline. The food is in contact with the spoon during the Pushing Phase but it fails to be cupped inside the spoon mouth in the end. There are various failure cases in this failure modes, e.g., the cashews are pushed with too much force so it jumps out of the space between the pusher and the spoon. Another failure case of this failure mode in the experiments is that some small or thin food items are stuck between the spoon and the pusher, not able to roll into the spoon mouth.

\noindent\textbf{Roll}: Roll is a failure mode of round foods that are easy to roll in the environment. During the Pushing Phase, either the pusher or the spoon exerts a force on the food item. The food builds momentum and may roll in the environment. If the round foods are in contact with each other, they are also likely to roll against each other and roll out of the scooping trajectory. Therefore, the scooper will fail to pick up the foods.

\noindent\textbf{Fall}: After the Pushing Phase, only part of the food is cupped into the spoon mouth. Therefore, when the spoon rotates in the Scooping Phase, the food may fall out of the spoon because of its unstable position. In general,  this type of failure mode is more common in Single baseline than other methods because the pusher in Single baseline acts like a static barrier without following the rotating motion of the spoon and fails to stabilize the food during the process.

\noindent\textbf{Break}: The failure mode of Break only happens to fragile foods because this category of foods are prone to deform and break while being squeezed under large forces from the spoon or the pusher.

\label{sec:failure}

\FloatBarrier
\section{Additional Rollouts}
\label{sec:qualitative}
\begin{figure}[!htb]
% \vspace{-0.5cm}
\includegraphics[width=\linewidth]{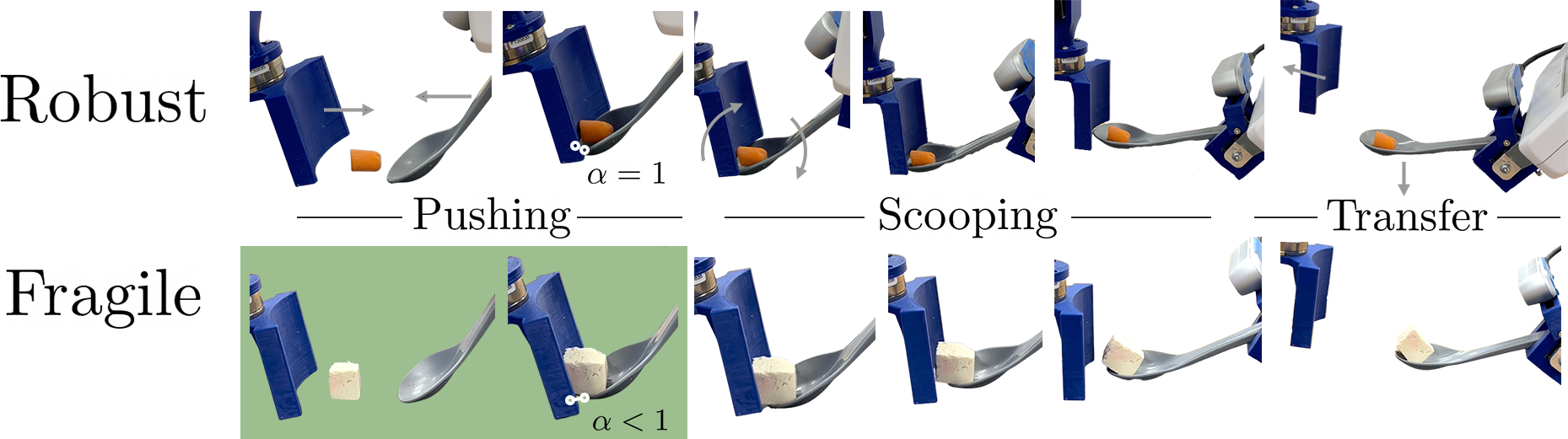}
\caption{\textbf{Full Scooping Rollouts}: Full rollout visualizations for scooping a robust food (carrot) and a fragile food (tofu) with adaptive pushing distance $\alpha$. Both foods employ the three phase bimanual scooping primitive, but differ in $\alpha$ choice. The carrot is a rigid food and uses and $\alpha$ for maximum stabilization during scooping, while the tofu is fragile and requires an $\alpha < 1$ to avoid breakage.}
\label{fig:rollouts}
\vspace{-0.3cm}
\end{figure}

\FloatBarrier

\section{Ablation Study of Stablizing Strategies}
\label{sec:ablation}
\revfour{Our approach \algabbr{} includes three stabilizing strategies, Angled Pushing, Adaptive Cupping and Pinning, to prevent potential failures while scooping food of various properties. To understand how these three stabilizing strategies impact performance, we compare \algabbr{} to three ablation baselines, \algabbr{} without Angled Pushing, \algabbr{} without Adaptive Cupping and \algabbr{} without Pinning, on five robust food classes. In \algabbr{} without Angled Pushing, the angle of the pusher $\theta$ is set to $0$ during the Pushing Phase. In \algabbr{} without Adaptive Cupping, the angled pusher and scooper move in sequence rather than simultaneously so that the pusher and the spoon cannot cup the food together. Lastly in \algabbr{} without Pinning, the pusher stays on the table during the Scooping phase rather than following the scooper up to pin the foods in the spoon. 

For the ablation experiments, we consider 5 different types of foods: grape, blueberry, macaroni, snow pea and cashews. These 5 classes cover a variety of visual and physical properties, including brittleness and geometry. We consider cashews to represent brittle foods and macaroni for compliant foods. In terms of geometry, we include varied sizes from small to large, e.g., blueberry, grape and snow pea, and a wide range of shapes, e.g., round grape and irregularly shaped snow pea. Table~\ref{table:ablation_studies} reports the success rate of scooping grape, blueberry, macaroni, snow pea, and cashews with ablations of \algabbr{}'s stabilizing strategies.

Based on the success rate, \algabbr{} without Angled Pushing has worse performance on cashews because two cashews become wedged between the vertical pusher and scooper. The Angled Pushing stabilizing strategy was designed with this failure mode in mind, and encourages food items to roll over the lip of the spoon and into the spoon bowl. \algabbr{} without Adaptive Cupping does not match the performance of our approach on 4 out of 5 food items, showing that Adaptive Cupping is critical for most food items. This is because in the Pushing phase, Adaptive Cupping cups the food to be centered with the spoon mouth and prevents the food rolling away. Lastly, \algabbr{} without Pinning also performs worse than \algabbr{} on all five food classes because when the pusher does not follow the scooper motion during the Scooping phase, food will easily fall out of the spoon bowl. }

% \begin{wrapfigure}{r}{0.55\textwidth}
\begin{table}[ht]
\centering
% \vspace{-1.2cm}
% \resizebox{\columnwidth}{!}{
 \begin{tabular}{ p{0.14\linewidth}  p{0.22\linewidth} p{0.23\linewidth}  p{0.13\linewidth}  p{0.08\linewidth} }

\multirow{2}{*}{\textbf{Food Type}} & \multicolumn{4}{c}{\textbf{Success Rate}}  \\
% \cline{2-4}
% \hline
% \multicolumn{1}{||c||}{} &  &   & A & B & C & D & E\\
 & w/o \emph{Angled Pushing} & w/o \emph{Adaptive Cupping} & w/o \emph{Pinning} & {\centering \emph{\algabbr{}}}  \\
% \multicolumn{2}{||c||}{}& \multicolumn{3}{c ||}{}& A & B & C & D & E\\
\hline
Grape & \textbf{5/5} & 3/5 & 3/5 & \textbf{5/5} \\  
% \hline
Blueberry & \textbf{5/5} & 3/5 & 4/5 & \textbf{5/5}\\ 
% \hline
Macaroni & \textbf{5/5} & 4/5 & 4/5 & \textbf{5/5} \\ 
% \hline
Snow Pea & \textbf{4/5} & \textbf{4/5} & 3/5 & \textbf{4/5} \\ 
% \hline
Cashews (2) & 5/10 & 6/10 & 6/10 & \textbf{7/10} \\ 
% \hline
\end{tabular}
\makeatletter
\def\@captype{table}
\makeatother
\caption{\textbf{Ablation Study Results:}
\revfour{\normalfont{ We report the per food item success rate over 5 trials of scooping robust foods with \algabbr{} strategies and three ablation baselines: \algabbr{} without Angled Pushing, \algabbr{} without Adaptive Cupping, and \algabbr{} without Pinning. As expected, we observe \algabbr{} achieves best overall performance across all 5 food classes in the table.
\algabbr{} without Angled Pushing fails on cashews due to their propensity to become wedged between the vertical pusher and scooper, while \algabbr{} without Adaptive Cupping only matches \algabbr{} performance on one out of five food items. \algabbr{} without Pinning performs worse than \algabbr{} on all five food items. These results suggest that the combination of all three bimanual stabilizing strategies (Angled Pusher, Cupping Motion, and Pinning Motion) are indispensable to the generalization and robustness of our method for scooping various food items.}
}}
% }}
% \vspace{-0.42cm}
\vspace{-0.3cm}
\label{table:ablation_studies}
\end{table}
% \end{wrapfigure}

% \todo{how to define the different level of properties}
% \begin{table*}[!htbp]
% \centering
% % \vspace{-0.2cm}
% % \resizebox{\columnwidth}{!}{
%  \begin{tabular}{ c | c | c | c| c  }
%  Food Type & Shape & Size & Thickness & Deformability \\ 
%  \hline
%  Broccoli & irregular & Medium & Medium\\
%  \hline
%  Grape  & round & Medium & Medium\\
%  \hline
%  Blueberry & round & Small & Medium\\
%  \hline
%  Pea & round &\\
%  \hline
%  Cashew & irregular\\
%  \hline
%  Goldfish & irregular\\
%  \hline
%  Carrot \\
%  \hline
%  Strawberry \\
%  \hline
%  Snowpea thin\\
%  \hline
%  Farfalle irregular\\
%  \hline
%  Macaroni \\
 
% \hline
% \end{tabular}
% \end{table*}

\section{Additional Experiments for Out of Distribution Food Items}
\revfour{In order to further test the generalization of our method, we extend our method to other unseen, scattered food items that also require a scooping mechanism in our daily life. The food tested in the Table~\ref{table:ood_exps} are cooked rice and couscous and they are grouped into a scoopable area on the workspace before scooping. Results in Table~\ref{table:ood_exps} suggests that \algabbr{} achieves comparable performance for cooked rice and couscous with human baseline. Both \emph{Human} and \algabbr{} have larger weight loss in couscous than rice, indicating that it is more difficult to scoop couscous because they are much smaller and more easily pushed out of the pushing and scooping path. We do not consider any grouping strategies for scattered foods, and leave this problem as an interesting direction for future work. }
 \begin{table*}[!htbp]
\centering
\vspace{-0.3cm}
% \resizebox{\columnwidth}{!}{
 \begin{tabular}{ c  c  c   }

\multirow{2}{*}{\textbf{Food Type}} & \multicolumn{2}{c}{\textbf{Avg. Weight Difference (\%)}} \\
% \multicolumn{1}{c}{Stabilizing parameter (mm)} \\ 
\multicolumn{1}{c}{}& \emph{\algabbr{}}  & \emph{Human}   \\
\hline
Rice (\algabbr{} OOD) & 10.339 & \textbf{8.423}\\ 
% \hline
Couscous (\algabbr{} OOD) & 31.666 & \textbf{21.594} \\
\end{tabular}

\caption{\textbf{OOD Food Results}: \revfour{We report the weight loss of food items after scooping as a percentage of the original food weight, averaged across 5 scooping trials. We compare our method \algabbr{} to human baselines over 2 unseen scattered food rice and couscous. We observe that both \algabbr{} and \emph{Human} could scoop a majority amount of scattered food on the plate.}}
\label{table:ood_exps}
\vspace{-0.3cm}
 \end{table*}

\section{Force Sensor Analysis}
\begin{figure} %[!htbp]
\centering
\vspace{-0.2cm}
\includegraphics[width=\linewidth]{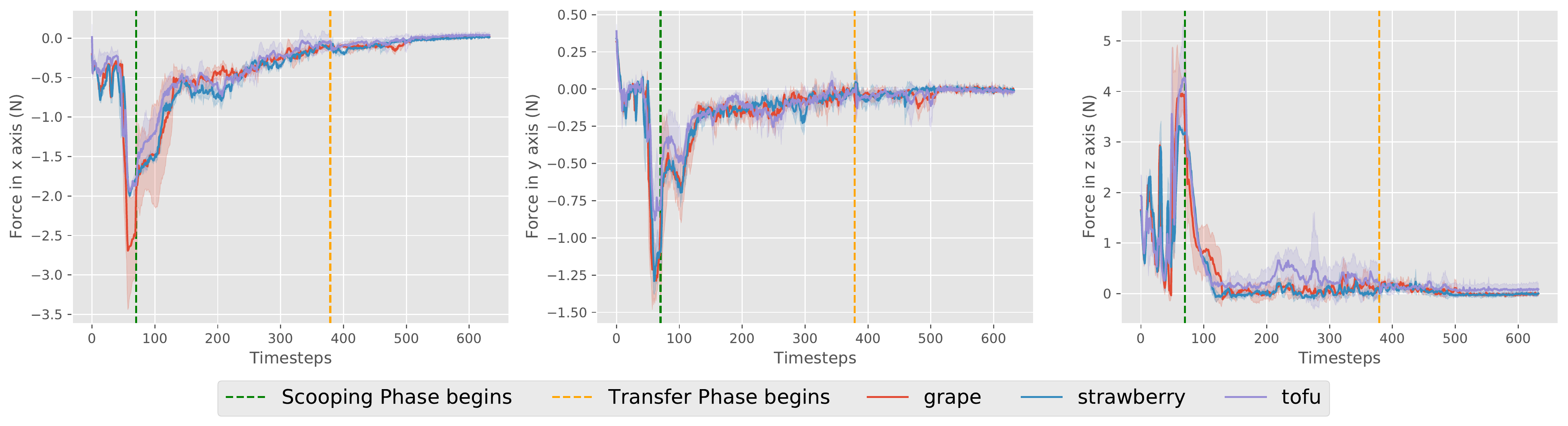}
\caption{\textbf{Sensored forces of 3 food items}:\revfour{ We report the per food item sensored force in the pusher over 3 trials of scooping foods along the x, y and z axes in three phases(Pushing, Scooping, Transfer). The confidence interval in the plot is 95\% and there is no obvious difference between the grape, strawberry and tofu. }}
\label{fig:sensor_force}
% \vspace{-0.73cm}
\end{figure}

\begin{figure} %[!htbp]
\centering
\vspace{-0.2cm}
\includegraphics[width=\linewidth]{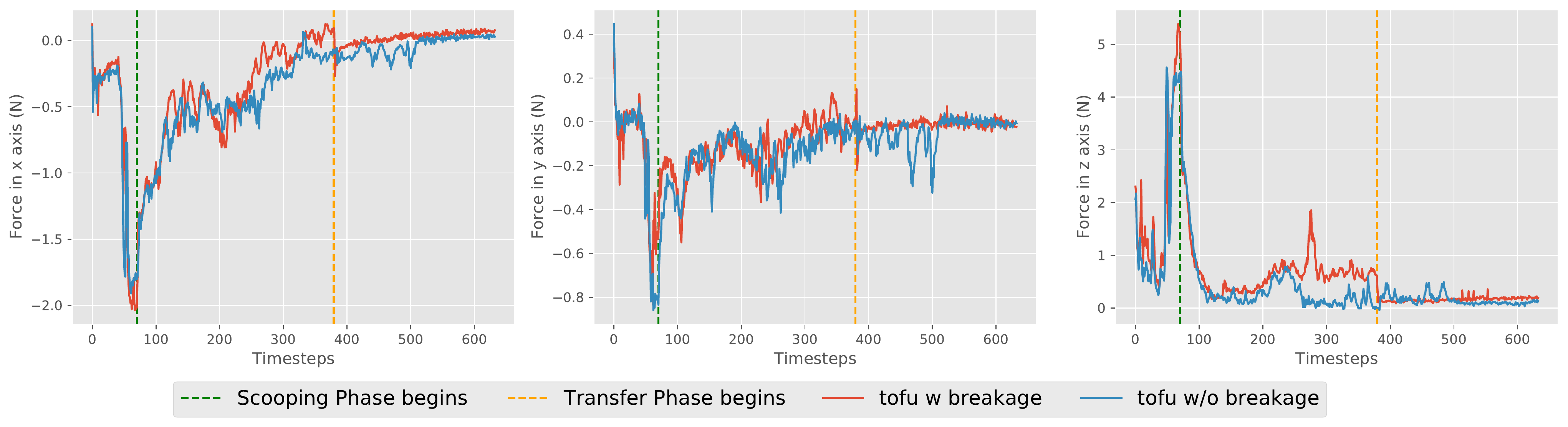}
\caption{\textbf{Sensored forces from tofu rollouts}:\revfour{ We report the sensored force in the pusher along the x, y and z axes in three phases (Pushing, Scooping, Transfer) for tofu rollouts. The breakage happens around the timestep of the green line where the transition from pushing to scooping starts. We cannot identify the breakage failures purely from the force torque sensing. }}
\label{fig:tofu_force}
% \vspace{-0.73cm}
\end{figure}

\revfour{Apart from the visual servoing system for bimanual scooping, we also consider using a force torque sensor mounted to the pusher. We conduct experiments to test if force sensor readings give additional information about the deformability of the food to distinguish between robust and fragile foods, as well as whether a fragile food is in a breakage-imminent state. We present the force readings along three axes during the three phases: Pushing, Scooping and Transfer, for three food items, including grape, strawberry and tofu.

Figure~\ref{fig:sensor_force} demonstrates that there are no distinct differences of forces between robust and fragile foods, suggesting that tactile information alone would not be sufficient for classifying food fragility. We also present the forces during two rollouts of scooping tofu. One of these rollouts has breakage at the transition between the Pushing and Scooping phases, while the other has no breakage throughout. The difference of forces in Figure~\ref{fig:tofu_force} is also too small to identify which one has the breakage, and when the breakage occurred. 

In conclusion, we find that the force-torque readings alone are too noisy to be used to replace the Risk Classifier and distinguish between robust and fragile food items. Even within fragile food scooping rollouts, we are unable to identify breakage failures from force-torque sensing alone, suggesting the need for a vision-based system for our Failure Classifier. Although past food acquisition works have used tactile sensing~\cite{gordon2021leveraging}, we hypothesize the multi-object interactions present in bimanual scooping, such as the pusher scraping against the plate, adds too much noise to the force readings to be used to identify food states. }

\end{document}